\documentclass[conference]{IEEEtran}
\IEEEoverridecommandlockouts

\usepackage{cite}
\usepackage{amsmath,amssymb,amsfonts}
\usepackage{graphicx}
\usepackage{textcomp}
\usepackage{xcolor}
\usepackage{nicefrac}
\usepackage{hyperref}
\usepackage{cleveref}
\usepackage{comment}
\usepackage{url}
\usepackage{verbatim}
\usepackage{booktabs}
\usepackage{tablefootnote}
\usepackage{threeparttable}
\usepackage{bm}
\usepackage{algorithm}
\usepackage{algpseudocode}
\usepackage{float}
\usepackage{subcaption}
\usepackage{tikz}
\usepackage{dblfloatfix}

\begin{document}

\title{Programmable Telescopic Soft Pneumatic Actuators for Deployable and Shape Morphing Soft Robots
}

\author{
\IEEEauthorblockN{Joel Kemp\IEEEauthorrefmark{2}\IEEEauthorrefmark{3}\IEEEauthorrefmark{1}, 
André Farinha\IEEEauthorrefmark{2}\IEEEauthorrefmark{1}, 
David Howard\IEEEauthorrefmark{2}, 
Krishna Manaswi Digumarti\IEEEauthorrefmark{3} and Josh Pinskier\IEEEauthorrefmark{2} }
\IEEEauthorblockA{\IEEEauthorrefmark{2} CSIRO Robotics, Australia | Contact: andre.farinha@csiro.au}
\IEEEauthorblockA{\IEEEauthorrefmark{3} Centre for Robotics, Queensland University of Technology, Australia}
\IEEEauthorblockA{\IEEEauthorrefmark{1} these authors contributed equally}
}


\maketitle

\begin{abstract} 
Soft Robotics presents a rich canvas for free-form and continuum devices capable of exerting forces in any direction and transforming between arbitrary configurations. However, there is no current way to tractably and directly exploit the design freedom due to the curse of dimensionality. Parameterisable sets of designs offer a pathway towards tractable, modular soft robotics that appropriately harness the behavioural freeform of soft structures to create rich embodied behaviours.
In this work, we present a parametrised class of soft actuators, Programmable Telescopic Soft Pneumatic Actuators (PTSPAs). PTSPAs expand axially on inflation for deployable structures and manipulation in challenging confined spaces. We introduce a parametric geometry generator to customise actuator models from high-level inputs, and explore the new design space through semi-automated experimentation and systematic exploration of key parameters.
Using it we characterise the actuators' extension/bending, expansion, and stiffness and reveal clear relationships between key design parameters and performance.
Finally we demonstrate the application of the actuators in a deployable soft quadruped whose legs deploy to walk, enabling automatic adaptation to confined spaces. PTSPAs present new design paradigm for deployable and shape morphing structures and wherever large length changes are required.
\end{abstract}

\section{Introduction}
Soft robotics offers a transformative potential to create both highly unconventional and truly lifelike robots. Freed from the constraints of traditional robot design, soft robotics unlocks a new world of possibilities in deployable, reconfigurable and shape adaptive robots, which exploit their form to function efficiently across environments \cite{KIM2013287,Baines2022,10637491}. The freeform and unconstrained design potential soft robotics presents an opportunity and a challenge, as the vast design space is too large to be meaningfully searched by even the best optimisation algorithms \cite{Pinskier2022}. Decomposing continuum robots into discrete modules offers a tractable alternative path, in which continuum modules can be designed independently then assembled into arbitrarily complex robots \cite{hexhidro,kriegman2019scalablesimtorealtransfersoft,Xu2025,Belke2023}. Modularity enables rapid robot design and an efficiently searchable design domain, with the only interconnections breaking the robot continuum. Whilst modularity typically involved homogenous modules, multiple distinct modules can also be incorporated, facilitating path generation and grammar-based robot evolution \cite{Zhao2020}. Where individual modules are parametrised and shape-programmable, complete, multi-environment robots can be efficiently generated through hierarchical optimisation \cite{Chand2021,Howard2019}. 

\begin{figure}[t]
    \centering
    \includegraphics[width=1.0\linewidth]{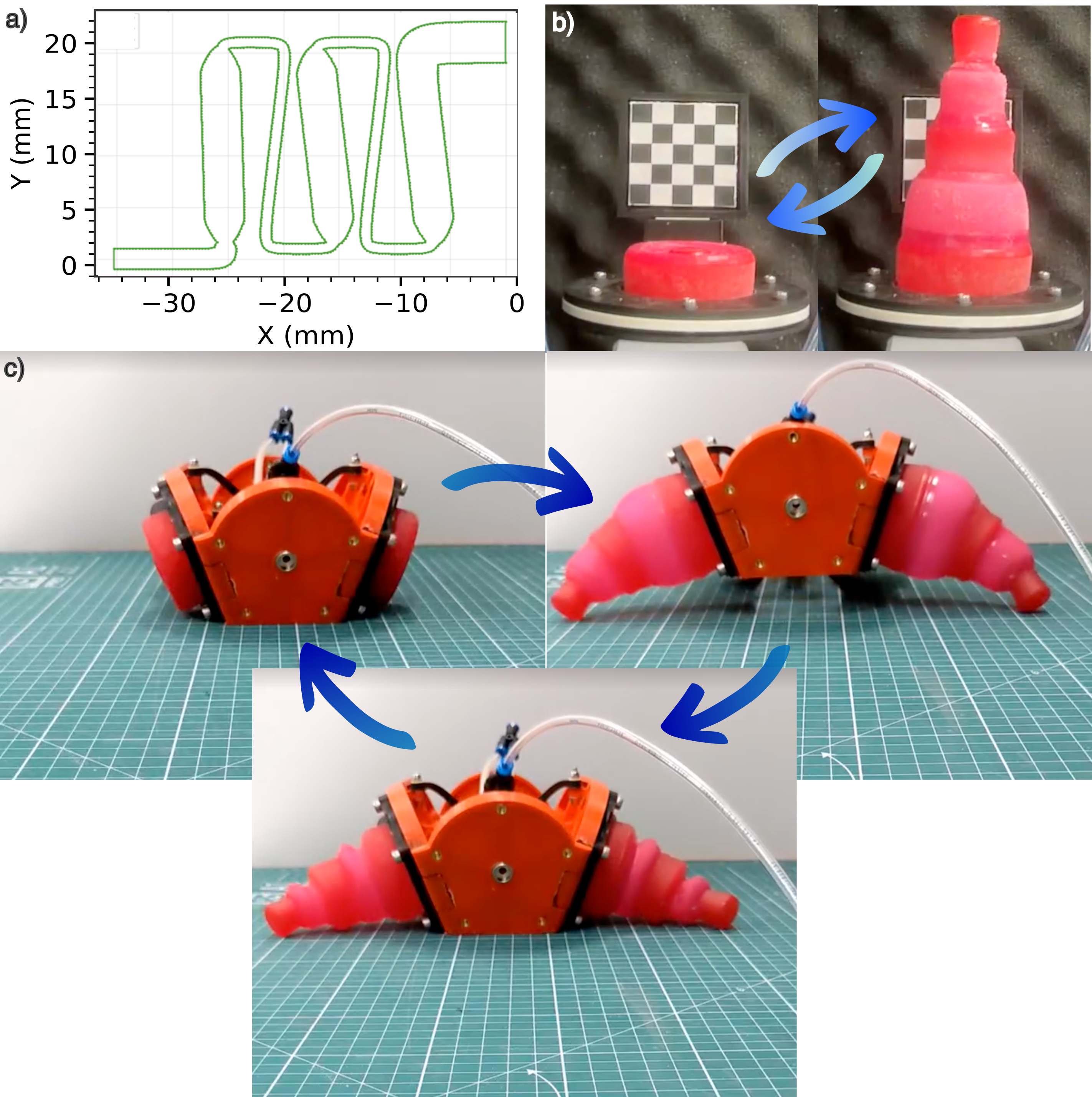}
    \caption{Programmable Soft Telescopic Actuator Overview: a) 2D curve produced by parametric actuator generator which is revolved to produce the 3D actuators. b) Resulting 3D Printed linear actuator at rest and inflated. c) Application of bending optimised actuators in soft quadruped. The actuators inflate with positive pressure, retain their shape at atmosphere and retract under vacuum}
    \label{fig:heroshot}
    \vspace{-12pt}
\end{figure}

\begin{table*}[t]
\begin{center}
\caption{Comparative Performance Metrics of Axial Soft Pneumatic Actuators}
\label{tab:actuators}
\begin{threeparttable}
\begin{tabular}{@{}lcccccl@{}}
\toprule
\textbf{Actuator Type} & \textbf{Max Force (N)} & \textbf{Response (s)} & \textbf{Expansion} & \textbf{Contraction} & \textbf{Ratio}\tnote{1} & \textbf{Source} \\
\midrule
\textbf{Origami: Zigzag} & 432 & $<$0.3 & Recovery weight & Vacuum & C: 90\% & \cite{Wood_origami} \\
\textbf{Origami: Kresling} & 6.4 & - & Fluid pressure & Vacuum & E–C: 130\% & \cite{tang2024bio} \\
\textbf{Origami: Yoshimura} & 50 & $<$2.5 & Fluid pressure & Vacuum & E: 72\% C:46\% & \cite{shen2021soft} \\
\midrule
\textbf{Telescopic} & 8.1 & 7–8 & Fluid pressure & Passive & E: 430\% & \cite{telescopic1}\\

\textbf{Telescopic (This Work)} & 5\tnote{2} & 1.5-3 & Fluid pressure & Vacuum & E: 650\% & \Cref{fig:pilot}.A\\
\midrule
\textbf{Bellows} & $>$30 & - & Passive & Vacuum & C: 52\% & \cite{qiu2024bellows} \\
\textbf{Bellows: Euglenoid} & - & ~60s & Fluid pressure & Vacuum & E:100 C: 50\% & \cite{digumarti2017euglenoid}\\
\bottomrule
\end{tabular}
\begin{tablenotes}
\item[1] Calculated as $(l_1 - l_0)/l_0$. Note that $l_1$ is taken as the length after a power stroke, so the maximum value in contraction cannot exceed 100\%, while under expansion there is theoretically no limit.
\item[2] Semi-automated tests capped at 5N
\end{tablenotes}
\end{threeparttable}
\end{center}
\vspace{-16pt}
\end{table*}

Existing soft robot modules have largely been based around bending, twisting and contracting actuators \cite{Zhang_2020}, particularly Soft pneumatic actuators (SPAs). SPA modules employ deformation modes including bending, twisting, axial extension/contraction, radial change and combinations of these to produce soft robots with conformable grasping and adaptive locomotion abilities \cite{gorissen2017elastic, polygerinos2017soft, xavier2022soft,Connolly2017,Katzschmann2018a,10.3389/fnbot.2019.00106}.
Of these, axial deformation is favoured to create thrust/elongation in artificial muscles, resulting in several classes of actuator with distinct mechanical principles, including origami-inspired, telescopic, and bellows actuators (\Cref{tab:actuators}). 
Origami-inspired SPAs use programmable folding to transfer forces directly without material stretching \cite{martinez2012elastomeric}. Bellows SPAs exploit corrugated soft structures that extend under positive pressure or contract via vacuum \cite{digumarti2017euglenoid, qiu2024bellows, joe2022review}.

\begin{figure*}[!b]
    \centering
    \includegraphics[width=\textwidth]{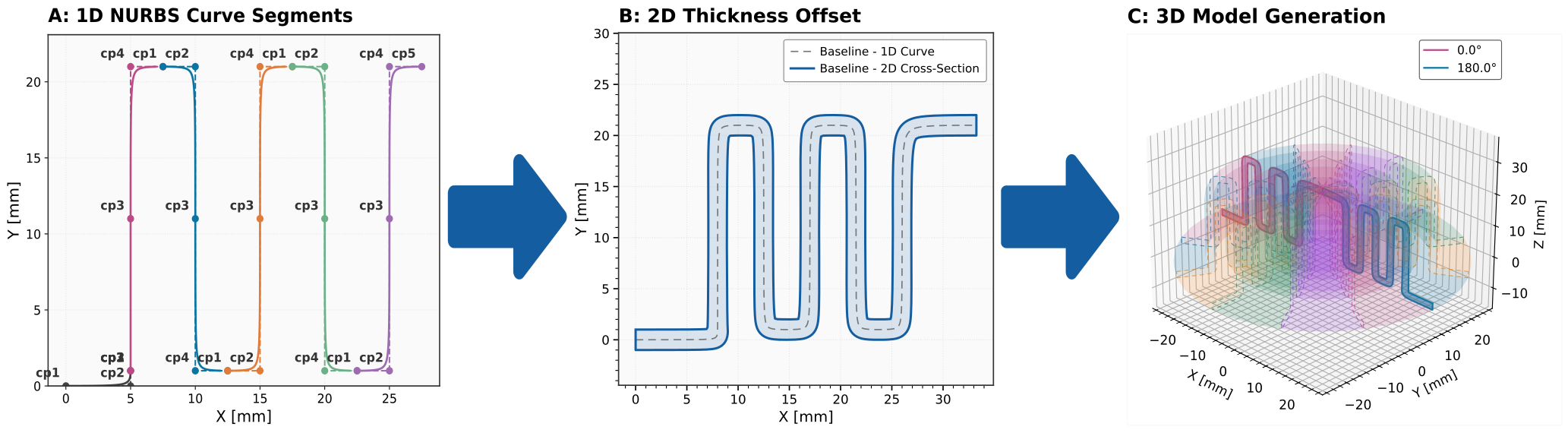}
    \caption{A: Visualisation of the NURBS segments and their union to construct the midline. B: Visualisation of the thickness offset applied to construct 2D cross-section. C: Visualisation of 3D construction via angular cross-sections. Shaded cross-sections are user-defined; dotted cross-sections are interpolated by the design framework $\theta_i = 30$.}
    \label{fig:framework_overview}
\end{figure*}

Compared to other classes of SPA, Telescopic SPAs are substantially under-explored, and have only been demonstrated only in soft grasping applications \cite{telescopic1,telescopic2, telescopic_thai}. Telescopic SPAs inflate pre-folded elastomeric chambers to extend in pre-defined shapes. When pressurized, concentric chambers inflate sequentially, creating a large extension from a compact base.
As the extension is created by material unfolding along, rather than perpendicular to, the extension axis, the axial and bending stiffness are largely decoupled, creating the possibility for programmable motion \cite{10.1145/3072959.3073673} and stiffness profiles, all within a deployable soft module.

To exploit these underexplored features of telescopic SPAs, we develop a framework for shape programmable telescopic soft pneumatic actuators (PTSPAs). We establish a flexible but tractable parameterised PTSPA generator and characterise the relationship between key design parameters in semi-automated testing, enabling the creation of bespoke, deployable PTSPA modules.
Our main contributions are: (i) the establishment of a new class of deployable shape programmable pneumatic actuators suitable for deployable structures, and optimisation and learning (ii) the development of a geometry generation pipeline to rapidly create bespoke modules, (iii) a structured exploration of the PTSPAs design domain using semi-automated testing to characterisation of PTSPA modules' performance and reveal design capabilities, and (iv) the demonstration of PTSPA capabilities in a multi-modal deployable 'Turtle-Roo' soft robot for unguided free and constrained space locomotion.

\section{Automated Design Framework}
\label{sec:design}

The geometry of telescopic soft actuators is defined by a set of parameters that describes a 1D curve (\Cref{ssec:oned}) and the construction of its corresponding 2D cross-section (\Cref{ssec:twod}). The 3D geometry is then constructed by smoothly connecting 2D cross-sections defined across multiple angular planes. This framework accommodates both axisymmetric designs (purely telescopic actuation) and asymmetric designs (bending during extension, \Cref{ssec:3D}).

\begin{figure*}[t]
    \centering
    \makebox[\textwidth][c]{%
        \hspace{-0.01\textwidth}%
        \includegraphics[width=1.05\textwidth]{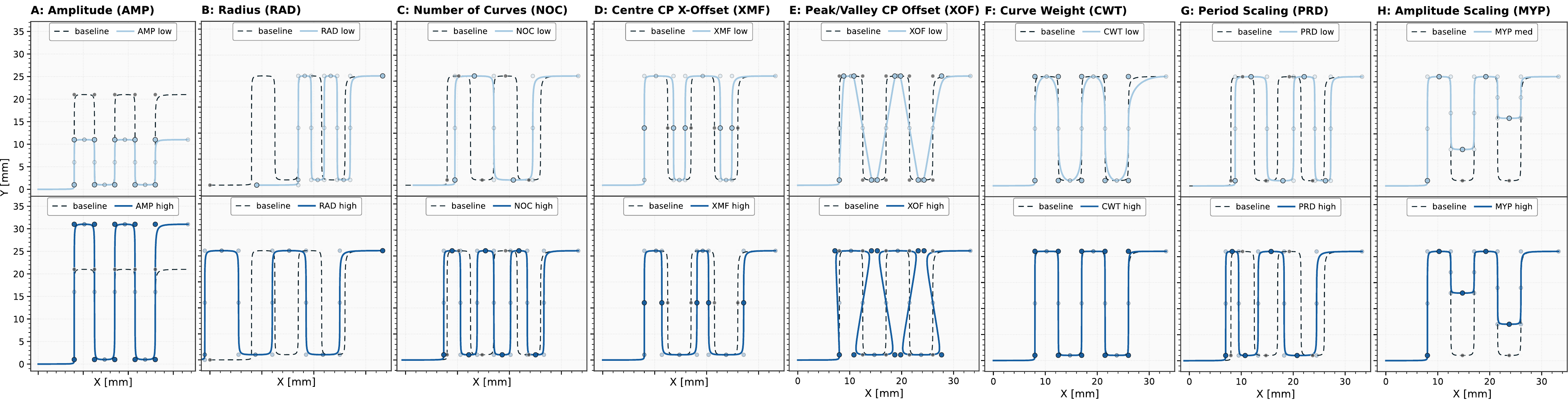}%
    }
    \caption{Geometric response of 1D input parameters.}
    \label{fig:1d_parameters}
        \vspace{-12pt}
\end{figure*}

\subsection{Midline Construction}
\label{ssec:oned}
The midline is controlled using Non-Uniform Rational B-splines (NURBS), selected for their local control properties and robust scaling behaviour. NURBS curves are defined as:

\begin{equation}
C(u) = \frac{\sum_{i=0}^n N_{i,p}(u)w_iP_i}{\sum_{i=0}^n N_{i,p}(u)w_i}
\end{equation}

where $N_{i,p}(u)$ are B-spline basis functions, $w_i$ are control point weights, and $P_i$ are control points. The knot vector $U=\{u_0, u_1, ... u_m\}$ maps control points to the local arc-length coordinate.

As shown in \Cref{fig:framework_overview}A, each B-spline segment consists of five control points and represents half of a curve period. The geometry of the curve is defined by the position of the control points $P_i$ and their weights $w_i$. While direct manipulation of all NURBS parameters would offer maximum flexibility, it would result in a prohibitively large and unintuitive set of parameters. Instead, we embed manufacturing and functional constraints through a reduced set of design parameters, whose geometric effects are illustrated in \Cref{fig:1d_parameters}.

The parameters \textit{Amplitude} and \textit{Radius} establish the overall scale, while all other parameters are normalized to these quantities, ensuring consistent scaling and robustness:

\begin{itemize}
    \item \textbf{Amplitude} (\Cref{fig:1d_parameters}A): Maximum vertical displacement of each NURBS segment.
    \item \textbf{Radius} (\Cref{fig:1d_parameters}B): Total span of midline.
    \item \textbf{Number of Curves} (\Cref{fig:1d_parameters}C): N NURBS segments
\end{itemize}

\noindent\paragraph*{Local Parameters (Periodic Symmetry)} These parameters define the curve profile's geometric characteristics by applying identical control-point settings across all NURBS segments:

\begin{itemize}
    \item \textbf{Center Control Point Offset} (\Cref{fig:1d_parameters}.E): Horizontal displacement of the central control point $CP_2$, which modulates internal volume distribution.
    \item \textbf{Peak/Valley Control Point Offset} (\Cref{fig:1d_parameters}.D): Horizontal offset of peak and valley control points $CP_1$ and $CP_3$, controlling bulging and arc-length.
    \item \textbf{Curve Weight (\Cref{fig:1d_parameters}.F)}: Curvature weight $w_i$ which tunes local curvature radius.
\end{itemize}

\noindent\paragraph*{Local Parameters (Periodic Asymmetry)} Asymmetry along the midline is introduced through parameters that vary control point positions across distinct NURBS segments:

\begin{itemize}
    \item \textbf{Period Scaling} (\Cref{fig:1d_parameters}.G): Relative horizontal span of each segment, distributing the total radius non-uniformly across sections.
    \item \textbf{Amplitude Scaling} (\Cref{fig:1d_parameters}.H): Scaling of vertical minima $CP_5$, controlling amplitude variation across periods.
\end{itemize}

\subsection{2D Cross-Section Generation}
\label{ssec:twod}

The 1D midline is expanded into a closed 2D cross-section (\Cref{fig:framework_overview}B) by assigning a local thickness $r(s)$ along the curve and applying a union-of-circles (UoC) operation. Each point $C(s_i)$ long the midline becomes the center of a circle with radius $r(s_i)$; their union produces the closed cross-section.

\begin{figure}[h]
    \centering
    \includegraphics[width=\columnwidth]{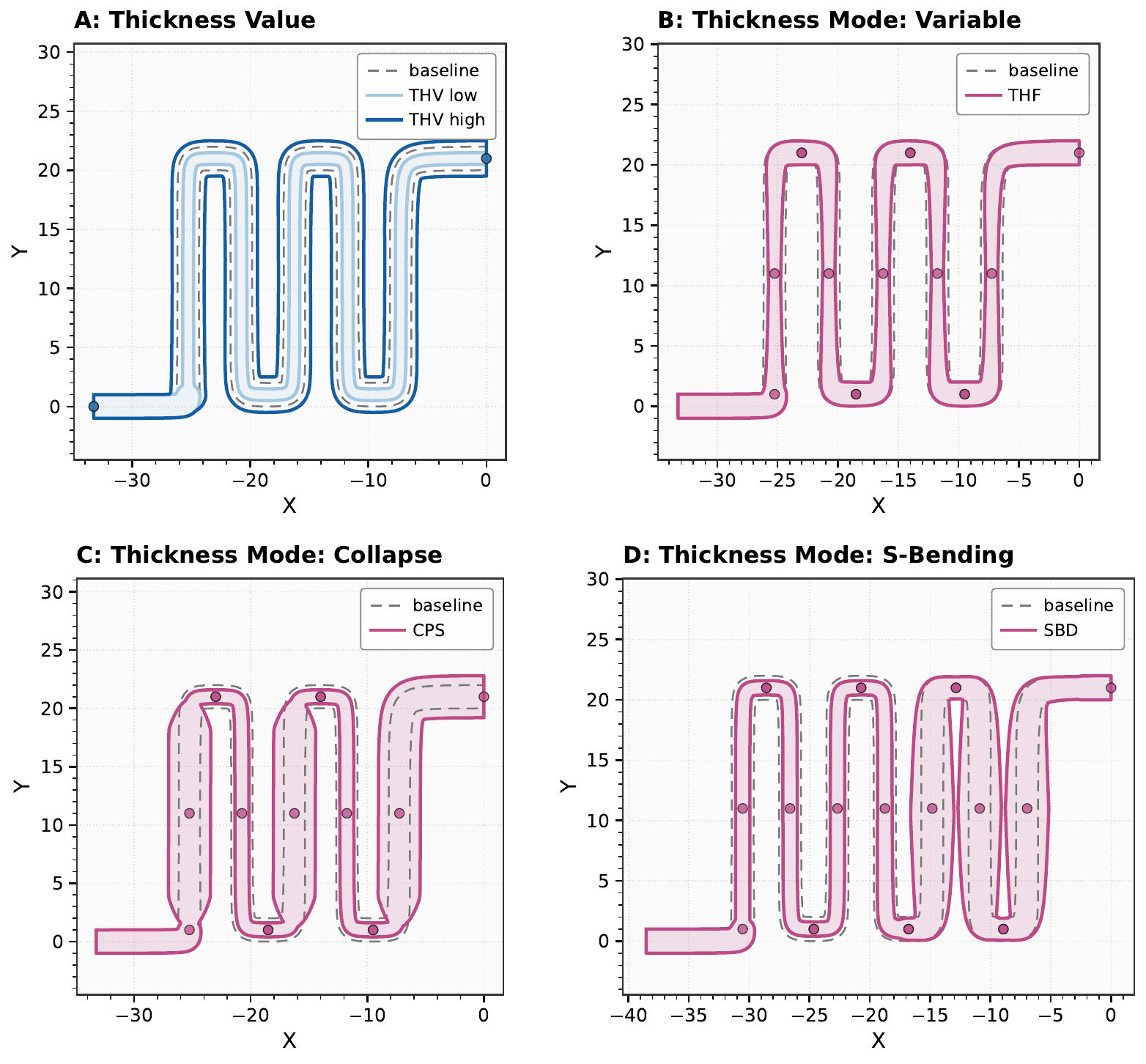}
    \caption{Visualisation of the geometric impacts of 2D input parameters}
    \label{fig:2d_parameters}
    \vspace{-12pt}
\end{figure}

This construction yields a thickness offset normal to the curve, which is robust to high curvature, avoids self-intersections, and permits thickness variation along the profile (\Cref{fig:2d_parameters}).

The 2D cross-section is controlled through three parameters:
\begin{itemize}
\item \textbf{Maximum Thickness}: Global thickness scale.
\item \textbf{Thickness Factors}: Local thickness scaling at control points.
\item \textbf{Thickness Mode}: Function defining thickness distribution along the curve.
\end{itemize}

Thickness factors are defined with respect to the control-point structure of NURBS segments. Each segment is tied to the control points (CP$_1$, CP$_3$, CP$_5$; see \Cref{fig:2d_parameters}B–D), with each point assigned a thickness factor. These factors are multiplied by the \textit{Maximum Thickness} to yield localized thickness values. The generator then maps each control point to its nearest sampled curve point, establishing a set of thickness anchor points along the curve.
The selected \textit{Thickness Mode} determines how thickness is interpolated between these anchors. The framework provides four modes:

\begin{itemize}
    \item \textbf{Constant Mode} \Cref{fig:2d_parameters}.A:  
    Applies uniform thickness to all midline samples: $r(s)=\textit{maximum thickness}$. 
    
    \item \textbf{Variable Mode} \Cref{fig:2d_parameters}.B:  
    The thickness factors $tf_1, tf_3, tf_5$ define the thickness at CP$_1$, CP$_3$, and CP$_5$ respectively, with $r(s_i)$ obtained via linear interpolation between these anchor points.
    
    \item \textbf{Collapsed Mode} \Cref{fig:2d_parameters}.C:  
    This mode applies the factors $tf_i$ to intercalated NURBS segments. In odd-count segments, instead of linear blending, thickness varies sharply after CP$_1$, and again before CP$_5$. Even-count segments maintain constant thickness throughout.
    
    \item \textbf{S-bend Mode} \Cref{fig:2d_parameters}.D:  
    Thickness is blended linearly, but different $tf_i$ are assigned to separate sets of NURBS segments.
\end{itemize}

\subsection{3D Model Generation}
\label{ssec:3D}

The final stage constructs the 3D actuator by lofting 2D cross-sections across angular planes $\theta_i$. While axisymmetric designs require a single cross-section; asymmetric designs specify sections at discrete angles, enabling deviations from purely axial expansion upon actuation.

Cross-section parameters can be provided at any subset of $\{\theta_1, \theta_2, \ldots, \theta_n\}$, and the generator interpolates between these planes and reconstructs intermediate sections via the 1D and 2D pipeline. Cross-sections are combined using CadQuery's \cite{CadQuery} loft operator (\Cref{fig:framework_overview}).C, with uniform arc-length resampling (\Cref{ssec:twod}) maintaining pointwise correspondence to prevent distortion.

\section{Experimental Methods}
\label{sec:exp}

This section outlines the experimental setup and methodologies used to evaluate the behaviour of the telescopic actuators generated by the proposed automated design framework. Our experiments aim to: (i) validate the performance of telescopic geometries output by the automated design framework, and (ii) explore the relationships between geometric parameters/characteristics and (deformation) behaviour of telescopic soft actuators. 

The design tool generates a 3D solid geometry of the actuator prototype which is readily printable and enables rapid prototyping of computationally generated designs. The prototypes are fabricated using a Stratasys J850 PolyJet printer, employing Agilus30 (shore 30A elastomer).

\begin{figure}[h]
    \centering
    \begin{subfigure}[t]{0.48\columnwidth}
        \begin{tikzpicture}
            \node[anchor=south west, inner sep=0] (image) 
                {\includegraphics[width=\columnwidth]{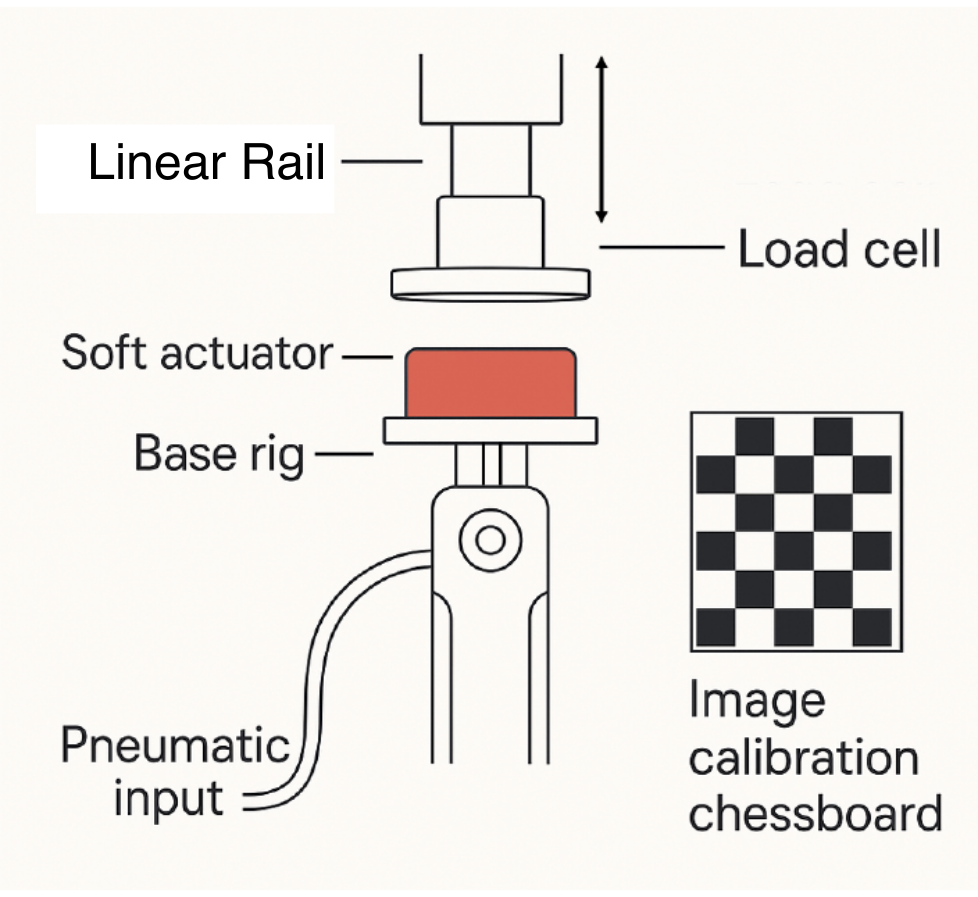}};
            \begin{scope}[x={(image.south east)}, y={(image.north west)}]
                \node[anchor=north west, font=\bfseries, fill=white, fill opacity=0.8, 
                      text opacity=1, inner sep=2pt] at (0.02,0.98) {A:};
            \end{scope}
        \end{tikzpicture}
    \end{subfigure}
    \hfill
    \begin{subfigure}[t]{0.50\columnwidth}
        \begin{tikzpicture}
            \node[anchor=south west, inner sep=0] (image) 
                {\includegraphics[width=\columnwidth]{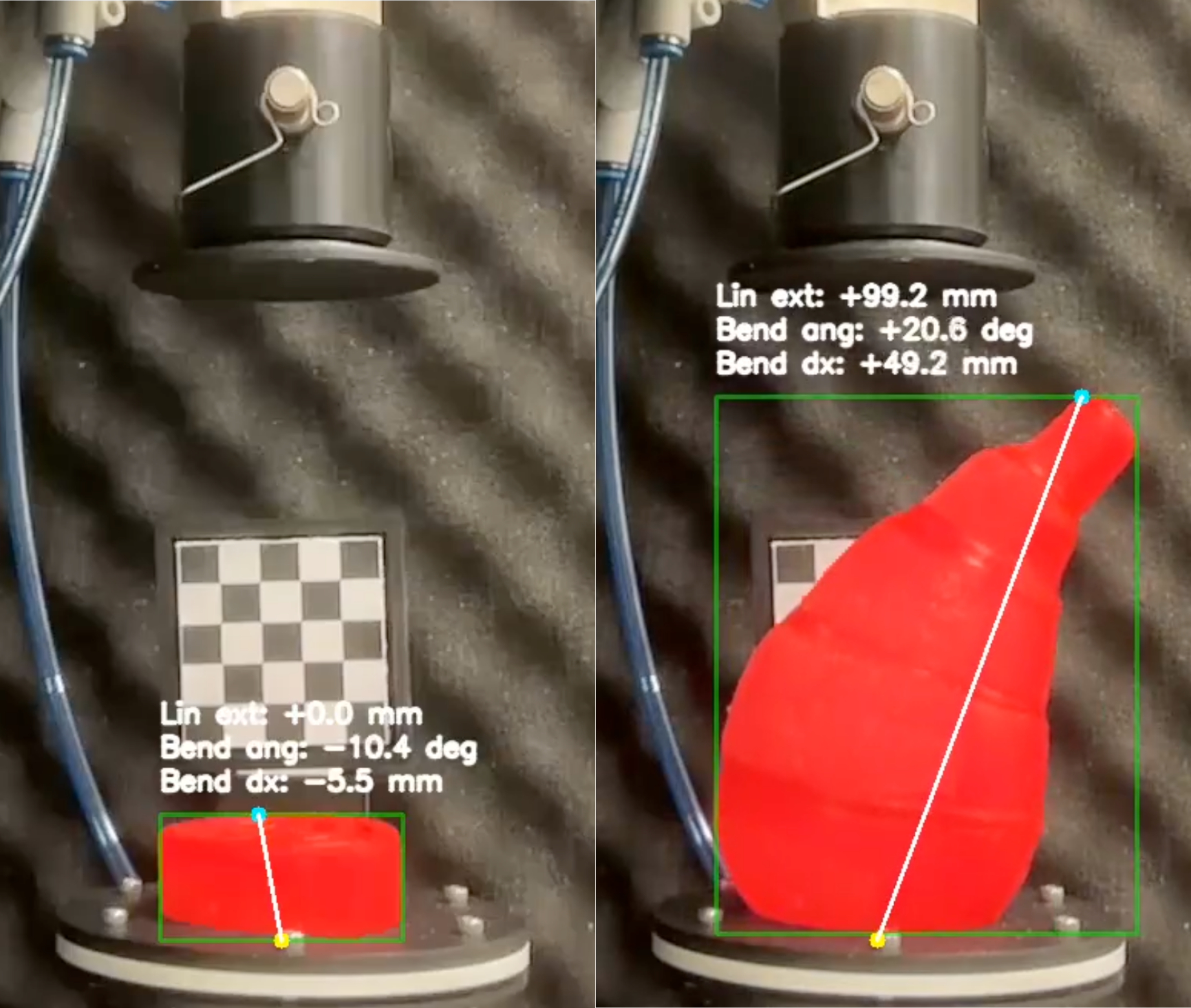}};
            \begin{scope}[x={(image.south east)}, y={(image.north west)}]
                \node[anchor=north west, font=\bfseries, fill=white, fill opacity=0.8, 
                      text opacity=1, inner sep=2pt] at (0.02,0.98) {B:};
            \end{scope}
        \end{tikzpicture}
    \end{subfigure}
    \caption{A) Automated experimental testing apparatus containing motorised linear rail, pneumatics infrastructure, and computer vision cameras. 
    B) Automated data extraction through computer vision processing}
    \label{fig:experimental}
    \vspace{-12pt}
\end{figure}

Testing was performed on an in-house automated linear testing platform (\Cref{fig:experimental}.a) equipped with integrated pneumatics, a load cell, and a linear stage (0.01 mm position tolerance). The platform is automated via a PC running ROS, allowing for further integration of a camera and data synchronisation. 


The experimental loop captures two distinct behaviours for every PTSPA prototype: (i) the deformation profile, and (ii) the stiffness characteristics.

All experimental videos were processed using an OpenCV pipeline to extract key deformation characteristics: (i) axial extension, (ii) radial expansion, and (iii) bending angle. 

Each image frame is first converted to HSV colour space and thresholded to isolate the actuator body. Morphological filtering removes noise, and the largest resulting connected region is taken as the actuator contour. From this contour a bounding box and enclosed area are computed, which serves as the basis for all kinematic measurements.

\begin{algorithm}[H]
\caption{Test Loop}
\begin{algorithmic}[1]
    \State \textbf{Phase 1: Model Search}
    \State \quad Move load-cell downward until a small contact force [0.1N] is detected.
    \State \quad Record the unactuated end-effector position $x_0$.
    \State \quad Move load cell to a safe distance height above $x_0$.

    \State \textbf{Phase 2: Pneumatic Actuation}
    \State \quad Begin video recording [Deformation Response].
    \State \quad Apply inflation for a fixed duration [3s].
    \State \quad End video recording [Deformation Response].

    \State \textbf{Phase 3: Stiffness Measurement}
    \State \quad Lower crosshead until a target contact-force threshold is reached [1N].
    \State \quad Repeat short inflation pulses for dynamic force capture.
    \State \quad Log force--displacement data for stiffness estimation.

    \State \textbf{Phase 4: Vacuum Cycle (Retraction Test)}
    \State \quad Begin video recording [Retraction Response]
    \State \quad Apply vacuum for a fixed duration [3s].
    \State \quad Vent to atmosphere and return to $x_\text{test}$.
    \State \quad End video recording [Retraction Response]

    \State \textbf{Phase 5: Reset}
    \State \quad Retract crosshead to a clearance height.
    \State \quad Stop logging and prepare for next prototype.
\end{algorithmic}
\end{algorithm}

Axial extension and radial expansion are computed relative to the lowest values observed during the run (at “rest” state). 

The axial extension and radial expansion are:
\begin{equation}
\Delta L = (h - h_{\min}),\,\quad
\Delta R = (w - w_{\min})
\end{equation}
where $h$ and $w$ are the bounding-box height and width, and $\alpha$ is the mm/px scaling factor calibrated using the chessboard reference. 

For bending experiments, a reference point is fixed at the lowest centre-point of the actuator, and an end-effector point is set at the highest point on the contour. The bending angle is computed from the vector joining the base to the tip:
\begin{equation}
\theta = \left|\arctan2(x_\mathrm{tip} - x_\mathrm{base},\, y_\mathrm{base} - y_\mathrm{tip})\right|.
\end{equation}
This derivation makes the assumption that the end-effector coincides with the highest point of the contour. This is useful for relative analysis but can generate unreliable data for SPA's that experience significant bending. Therefore, bending kinematics are manually verified.

\section{Results}
To validate the proposed parametric design framework and explore the design space detailed in \Cref{sec:design}, a series of 28 prototypes is generated, spanning a range of geometric configurations and subjected to the testing protocol detailed in \Cref{sec:exp}. The resulting deformation and contact force responses are analysed with respect to key performance characteristics, and the impact of design parameters on shape control is explored. 

\subsection{Design space exploration}

An initial set of 15 designs is generated by varying selected design parameters as compiled in \Cref{tab:factors}. While a sample of 3 prototypes (inclusive of baseline) is insufficient to represent the non-convex performance landscapes typically associated with soft actuators, it forms a valuable initial exploration to identify relevant design parameters. 

\begin{table}[h]
    \centering
    \caption{Summary of design parameters varied. Each parameter’s effect is illustrated in the figures referenced in the right column.}
    \label{tab:factors}
    \begin{tabular}{@{}llll@{}}
        \toprule
        \textbf{Parameter} & \textbf{Values (Low)} & \textbf{Values (High)} & \textbf{Figure} \\ 
        \midrule
        \textbf{AMP} Amplitude          & 10                     & 30                     & \Cref{fig:1d_parameters}.A \\
        \textbf{PRD} Period Scaling     & [1.5, 1, 0.5]          & [0.5, 1, 1.5]          & \Cref{fig:1d_parameters}.G \\
        \textbf{XOF} X-Offset Factor    & -0.75                  & 0.75                   & \Cref{fig:1d_parameters}.E \\
        \textbf{XMF} X-Centre Factor       & -0.5                   & 0.5                    & \Cref{fig:1d_parameters}.D \\
        \textbf{CWT} Curve Weight       & 1                      & 10                     & \Cref{fig:1d_parameters}.F \\
        \textbf{THF} Thickness Factor   & [1, 0.5, 1]            & [0.5, 1, 0.5]          & \Cref{fig:2d_parameters}.B \\
        \textbf{THV} Thickness Value    & 0.5                    & 1.5                    & \Cref{fig:2d_parameters}.A \\
        \bottomrule
    \end{tabular}
\end{table}

\begin{figure}[h]
    \centering
    \includegraphics[width=0.85\columnwidth]{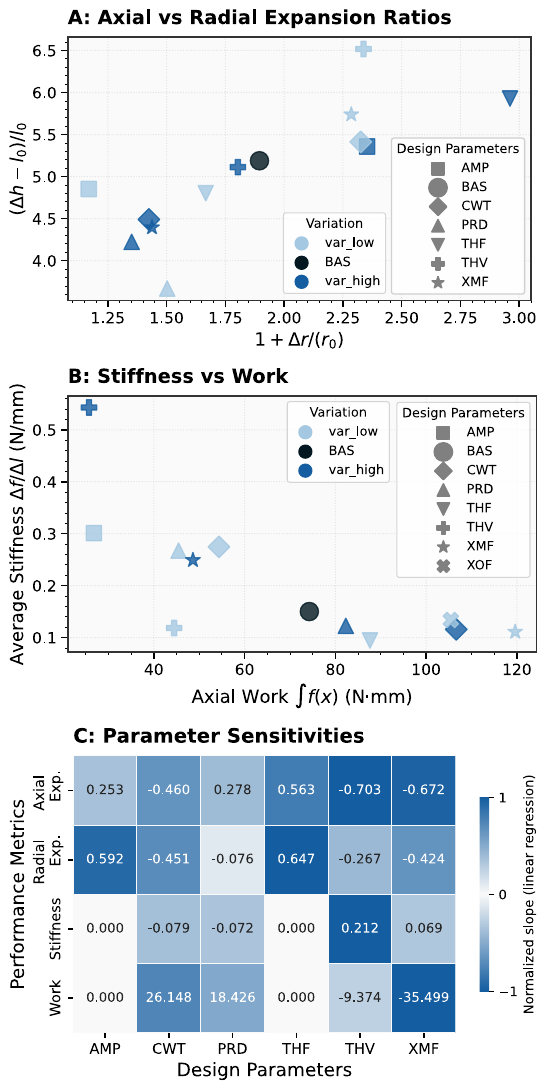}
    \caption{Actuator responses of the design-space exploration set in \Cref{tab:factors}, with all prototypes shown as variations from the baseline geometry "BAS". A: Axial expansion relative to initial actuator length vs the radial expansion relative to initial radius. B: Average stiffness vs axial work during axial contact interaction. C: Sensitivity of performance metrics to key design parameters.}
    \label{fig:pilot}
\end{figure}

\Cref{fig:pilot} condenses the experimental results obtained from this initial dataset and characterises the performance of each actuator in terms of the following normalised parameters: axial and radial expansion, and the average stiffness and work performed during an axial contact interaction. 

As shown in \Cref{fig:pilot}.A, the greatest axial expansion ratio obtained (value of 6.5) corresponds to the design with reduced wall thickness due to a pronounced second stage of inflation after telescopic expansion occurs. Another interesting observation can be made by varying the amplitude of the designs: while the relative axial expansion remains relatively unchanged, amplitude has a great impact on the axial expansion. In fact, the reduced amplitude design has the greatest axial/radial expansion ratio at 3.6.

\begin{figure*}[t]
    \centering
    \makebox[\textwidth][c]{%
        \hspace{-0.03\textwidth}%
        \includegraphics[width=1.05\textwidth]{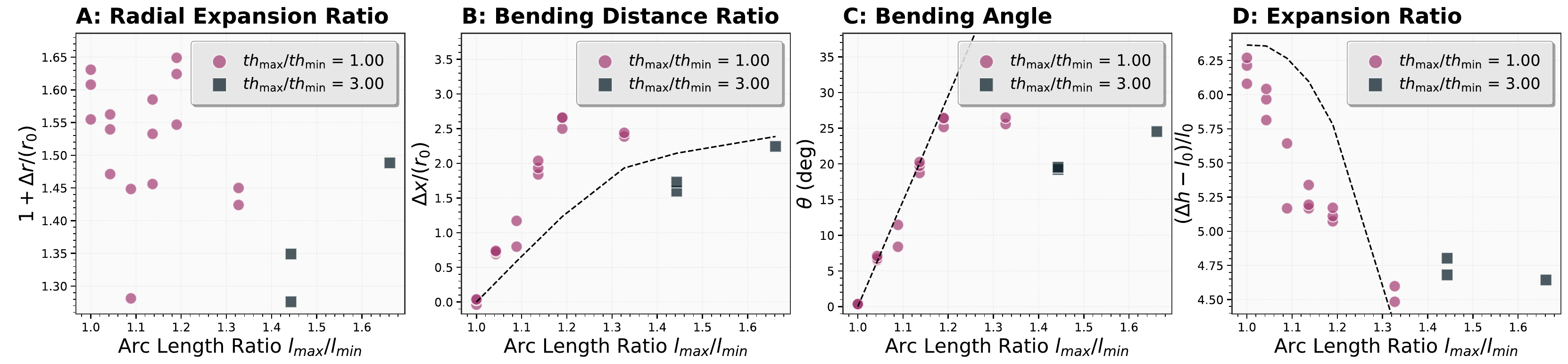}%
    }
    \caption{Bending of PTSPAs as function of the arc length ratio of diametrically opposed cross-sections. A: Relative radial expansion ratios; B: Transverse displacement of actuator tip normalised by initial radius; C: Bending angle measured from the centre of actuator to to its tip; D: Axial expansion ratio normalised by initial height; Trend-lines: Inextensible arc-length model of bending behaviour. }
    \label{fig:bending}
        \vspace{-12pt}
\end{figure*}

Unsurprisingly, this same design exhibits the second highest stiffness and lowest work (\Cref{fig:pilot}.B), simply due to its short expanded length and higher thickness to length ratio. It is also shown in that the designs measuring the highest axial work correspond to variations in control point offsets and weights, hinting at some more complex interplay in parameters than the ones previously highlighted. 

\Cref{fig:pilot}.C highlights the sensitivities of the highlighted performance metrics to relevant parameters calculated as the slope of fit linear models. It is shown that variation in flat thickness values, thickness variation and, surprisingly, the centre control point offset dominate all performance metrics, with a fourth relevant contribution of amplitude to the radial expansion. 

\begin{figure}[h]
    \centering
    \includegraphics[width=1\columnwidth]{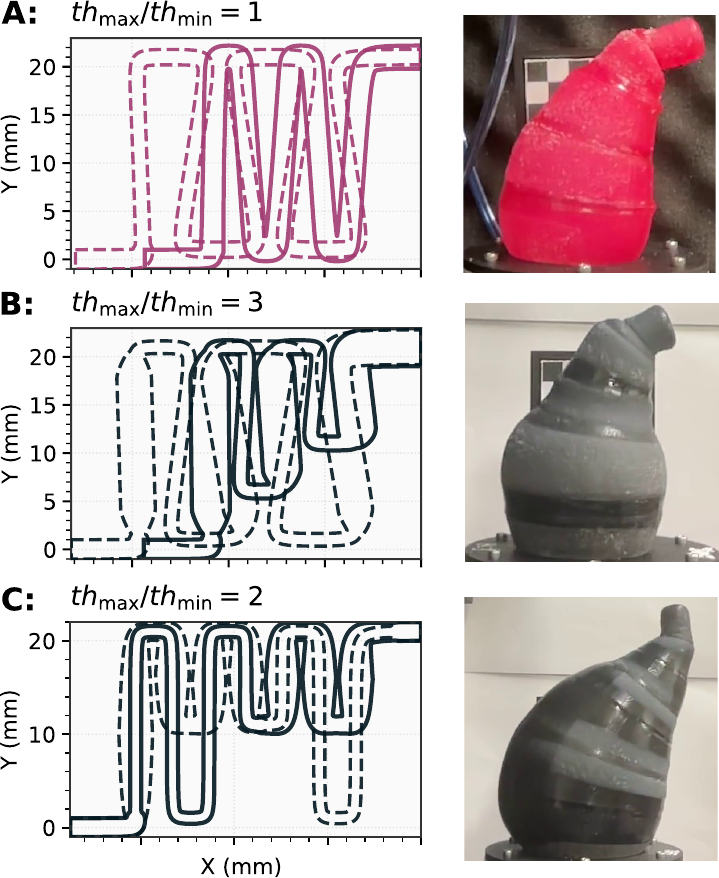}
    \caption{Curvature programming of PTSPAs via cross-section control. Left column: two diametrically opposed cross-sections which are interpolated to revolve the cross-section as in \Cref{ssec:3D}. A: Bending profile at constant thickness; B: Bending profile at variable thickness; C: S-shaped actuator profile. }
    \label{fig:shapes}
        \vspace{-12pt}
\end{figure}

\subsection{Shape Programming}

As discussed in \Cref{ssec:3D}, the cross-section of PTSPAs can be varied along the range of the revolve operation. In this work we have explored the use of the extrema control point offsets (see \Cref{fig:1d_parameters}.E), amplitude scaling (see \Cref{fig:1d_parameters}.H) and variation in thickness distribution (see \Cref{fig:2d_parameters}.D). As shown in \Cref{fig:shapes}, PTSPAs can be designed to steer their telescoping motion away from the purely axial direction. Curvature arises from introducing circumferential variation in the cross-section, thereby breaking axisymmetry.
This non-uniformity can be achieved either by varying the midline arc length or the wall thickness along the angular direction. A locally thicker wall causes the actuator to bend towards that region, whereas a longer cross-section segment induces bending away from it. As illustrated in \Cref{fig:shapes}C, combining these two effects out of phase produces an overall S-shaped profile.

Although both wall thickness and arc length can be used to generate curvature, it is preferable to leave the thickness unconstrained, given its dominant influence on a variety of other behaviours (see \Cref{fig:pilot}).
Arc length can be modified through several geometric parameters, such as control point offsets and weights, amplitude scaling, and radius (see \Cref{fig:2d_parameters}B, E, F, and H). \Cref{fig:shapes}A illustrates a design that relies exclusively on peak and valley control point offsets to achieve bending, whereas \Cref{fig:shapes}B combines both offsets and amplitude scaling.
Despite the differences in bending mechanism, \Cref{fig:bending}B and C show that all designs follow a similar trend: the bending magnitude plateaus beyond a certain arc-length ratio. This behaviour, however, does not extend to the axial and radial expansion ratios, where the variation in thickness used in design B comes into play.

Interestingly, the bending behaviour can be modelled to some extent using an inextensible model that assumes the actuator’s final shape to approximate a tilted cone, whose cross-section is described by \Cref{eq:bend}, where $s_1$ and $s_0$ are the arc lengths of diametrically opposed sides of the actuator, $h$ is the actuator’s axial projection, $r$ is the base radius, 
and $x$ is the reach in transverse direction measured from the base centre. The system can be manipulated algebraically to obtain bending distances, angles, and axial expansion values. As shown in \Cref{fig:bending}, this very simple model is a good predictor of the bending angle before the observed plateau, and also follows the trend in axial expansion reasonably well, though those results are dependent on the material strain and thus inflation pressure.

\begin{equation}
\begin{aligned}
    s_1^2 &= h^2 + (x + r)^2, \\
    s_0^2 &= h^2 + (x - r)^2,
\end{aligned}
\label{eq:bend}
\end{equation}

\begin{figure*}[b]
    \centering
    \includegraphics[width=1\textwidth]{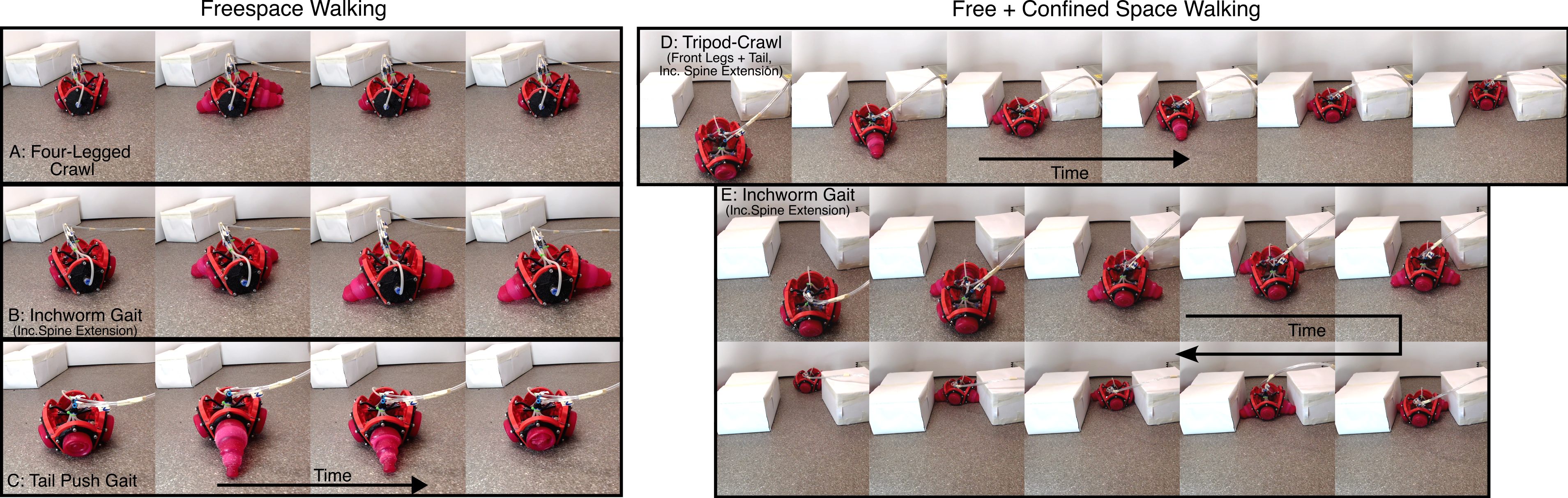}
    \addtocounter{figure}{1}
    \caption{Compilation of 'Turtle-Roo' Locomotion. A: 
    A: Four legged crawl, all legs are simultaneously actuated. B: Inchworm Gait, the spine is extended between front and rear leg cycles. C: Tail Push, in confined spaces forward thrust can be generated solely by pushing of the tail. D: Hybrid Tripod-Crawl Gait for narrow openings; the front legs, tail and spine are actuated in and inchworm-like cycle. E: Inchworm gait navigating a narrow opening}
    \label{fig:turtle compilation}
        \vspace{-12pt}
\end{figure*}

\begin{figure}[t]
    \centering
    \includegraphics[width=1\columnwidth]{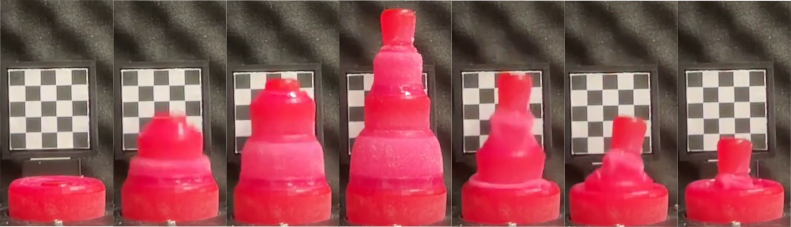}
    \addtocounter{figure}{-2}
    \caption{Actuation cycle of a PTSPA}
    \label{fig:retraction}
    \vspace{-12pt}
\end{figure}

\subsection{Actuation}

As documented in \Cref{tab:actuators}, soft pneumatic SPAs reported in prior literature typically extend under pressure and passively retract to a compact state upon depressurization. However, this passive retraction mechanism is dependent on materials with low hysteresis. A criterion not met by the photopolymers used in this work for rapid prototyping. Consequently, vacuum actuation is required to achieve reliable 2-way actuation. This introduces a critical challenge: SPAs undergo radial collapse when subjected to vacuum, compromising functionality and durability.

To mitigate radial collapse, we manipulate the actuator's cross-sectional geometry by introducing intercalating sections of varying thickness, as illustrated in \Cref{fig:2d_parameters}. These locally reinforced regions increase radial stiffness and enable controlled retraction under vacuum, as shown in \Cref{fig:retraction}. Whilst passive retraction is beneficial in many circumstances, our bistable PTSPAs can remain extended passively both when pressured and at atmosphere, enabling additional capabilities in both stiffness and friction adaptation.

\subsection{Demonstration of PTSPA in Quadruped Robot}
We demonstrate the benefits of the PTSPA through a deployable soft 'Turtle-Roo'. Turtle-Roo contains 6 actuators, 4 bending actuators as the legs and two linear extension actuators in the tail and as an extensible spine. Using its 4 legs (bending actuators) plus extendable spine and tail (linear extension), Turtle-Roo can exploit several gaits to traverse free and confined spaced using simple harmonic motions (
\Cref{fig:turtle compilation})

In free space, Turtle-Roo can elongate its spine to take large steps in an inchworm pattern, moving ~0.5 body lengths per cycle. On rough terrain, it can switch to slower but more stable gaits like the four-legged crawl, and in confined spaces it can use its tail to push itself forwards.

In confined spaces the deployability and softness of the actuators enables morphological trajectory computation, such that the robot can pass through the narrow opening without any sensing or feedback \Cref{fig:turtle compilation}(D-E). The deployable arms will naturally push against the cavity walls such that Turtle-Roo automatically steers towards the exit.

\section{Discussion and Conclusion}
This paper presents Programmable Soft Telescopic Pneumatic Actuators, a new design paradigm for deployable and shape morphing soft robots. Beyond simply extending and contracting, our geometry generation pipeline can rapidly create new soft modules with customised elongation, stiffness, shape, and radial expansion. We show extensions of upto 650\% can be reached with a 3 layered linear extension actuator. The modules can be connected serially or in parallel to produce complex deployable structures, including Turtle-Roo, our 4.5 legged (four legs and tail), extensible, deployable robot. It combines elements of quadrupeds, kangaroos and insects to produce multimodal locomotion suitable for freespace, rough terrain and narrow openings; and exploits its embodied intelligence to navigate crevasses without external feedback.

We believe PTSPAs can be applied broadly in deployable and shape morphing structures, where large length changes are required, including fully soft and hybrid soft-rigid robots. In addition to the benefits highlighted in our demonstrator robot, PTSPA 'intelligence' can be further expanded through precise pressure control and feedback, which can both control the deployment of actuator layers and detect contact for automated gait adaptation.
However, further work is required to increase the lifespan and responsiveness of the PTSPAs. Prototype polyjet printed designs have a lifespan of less than 100 cycles and are prone to bursting under pressure.

\bibliographystyle{vancouver}
\bibliography{refs}

\end{document}